\documentclass[letterpaper, 10 pt, conference]{ieeeconf}  

\IEEEoverridecommandlockouts                              

\usepackage{graphics} 
\usepackage{epsfig} 
\usepackage{mathptmx} 
\usepackage{times} 
\usepackage{amsmath} 
\usepackage{amssymb}  
\usepackage{colortbl}
\usepackage{xcolor}

\title{\LARGE \bf
Hybrid Continuum-Eversion Robot: Precise Navigation and Decontamination in Nuclear Environments using Vine Robot
}

\author{Mohammed Al-Dubooni\textsuperscript{1}, Cuebong Wong\textsuperscript{2}, and Kaspar Althoefer\textsuperscript{1}, Senior Member, IEEE
\thanks{This work has been submitted to the IEEE for possible publication. Copyright may be transferred without notice, after which this version may no longer be accessible.}
\thanks{*Purpose of open access, the author(s) has applied a Creative Commons Attribution (CC BY) license to any Accepted Manuscript version arising. }%
\thanks{\textsuperscript{1}Authors are with the Centre for Advanced Robotics @ Queen Mary, School of Engineering and Materials Science.}%
\thanks{\textsuperscript{2}Authors are with the National Nuclear Laboratory.}%
\thanks{Research is funded by the Nuclear Decommissioning Authority PhD Bursary programme.}%
}

\begin{document}

\maketitle
\thispagestyle{empty}
\pagestyle{empty}

\begin{abstract}
Soft growing vine robots show great potential for navigation and decontamination tasks in the nuclear industry. This paper introduces a novel hybrid continuum-eversion robot designed to address certain challenges in relation to navigating and operating within pipe networks and enclosed remote vessels. The hybrid robot combines the flexibility of a soft eversion robot with the precision of a continuum robot at its tip, allowing for controlled steering and movement in hard to access and/or complex environments. The design enables the delivery of sensors, liquids, and aerosols to remote areas, supporting remote decontamination activities. 

This paper outlines the design and construction of the robot and the methods by which it achieves selective steering. We also include a comprehensive review of current related work in eversion robotics, as well as other steering devices and actuators currently under research, which underpin this novel active steering approach. This is followed by an experimental evaluation that demonstrates the robot’s real-world capabilities in delivering liquids and aerosols to remote locations. The experiments reveal successful outcomes, with over 95\% success in precision spraying tests. The paper concludes by discussing future work alongside limitations in the current design, ultimately showcasing its potential as a solution for remote decontamination operations in the nuclear industry.
\end{abstract}

\section{Introduction}
Soft eversion robotics is now a rapidly growing field that has been driven by the need to develop devices that can operate safely and effectively in areas that are hard to access using traditional rigid bodied robots.  Soft compliant robots can navigate through tight spaces, making them ideal for applications in fields such as healthcare \cite{c1}, \cite{c18}, \cite{c19} manufacturing, and exploration \cite{c2}, \cite{c3}. Eversion robots are a specialized type of soft robot designed to invert or unfold \cite{c4}, \cite{c5}, \cite{c20}, in turn, exhibiting a unique form of locomotion.

The primary characteristic of eversion robots is their ability to undergo continuous transformation, inverting their structure to adapt to various environments \cite{c21} or perform specific tasks when pressurised by an air supply \cite{c6} . This design is particularly advantageous when navigating through constrained spaces. Most eversion robots are constructed from nylon ripstop fabric due to its useful combination of properties, including flexibility, durability, and tear resistance, which allow for a seamless inversion process while withstanding the mechanical stresses of repeated deformations \cite{c7}. 

Looking at the nuclear industry \cite{c8} we identify three significant operational challenges associated with nuclear decommissioning, these being  navigation, sensing and sealing. The first of these involves access into and operation within different sized and different shaped spaces, ranging from chemical piping to AC ducting and underground crawl spaces. The other two challenges go hand in hand; the delivery of sensors such as radiation sensors \cite{c9}, \cite{c10}, \cite{c11} and the delivery of a liquid such as an “antidote” to  potentially harmful matter,  a sticky substance like fix-it, or an expanding foam used to immobilise  potentially hazardous materials (such as radioactive contaminants in pipes). 

\begin{figure}[t]
  \centering
  \includegraphics[width=1\columnwidth]{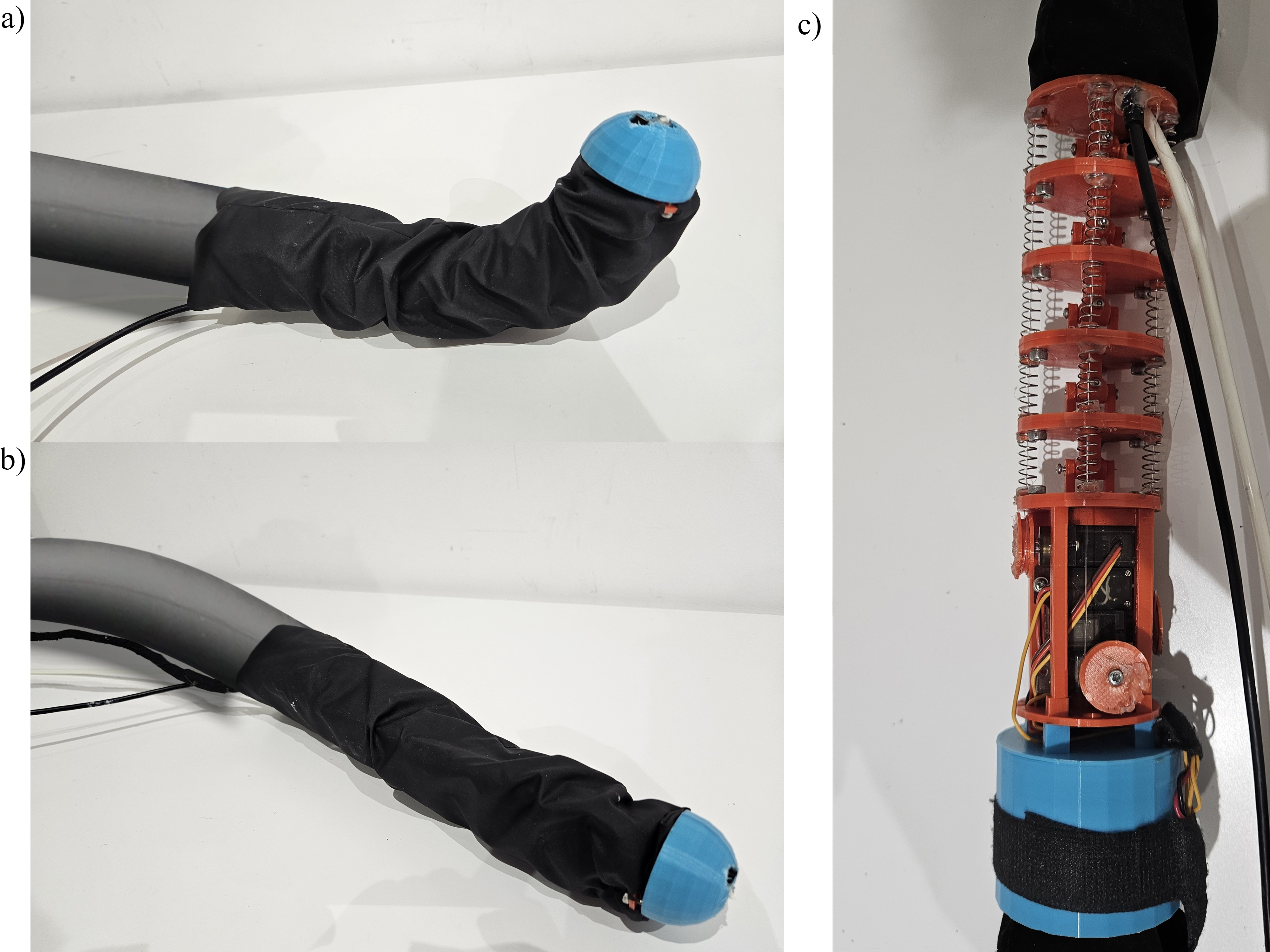}
  \caption{Innovative Hybrid Continuum-Eversion Robot: Selective tip steering enabled by a rigid-component robotic structure, demonstrated by a) and b). Subfigure c) shows the tip design without the nylon sleeve covering.}
  \label{fig1}
\end{figure}

\begin{figure}[b]
  \centering
  \includegraphics[width=0.6\columnwidth]{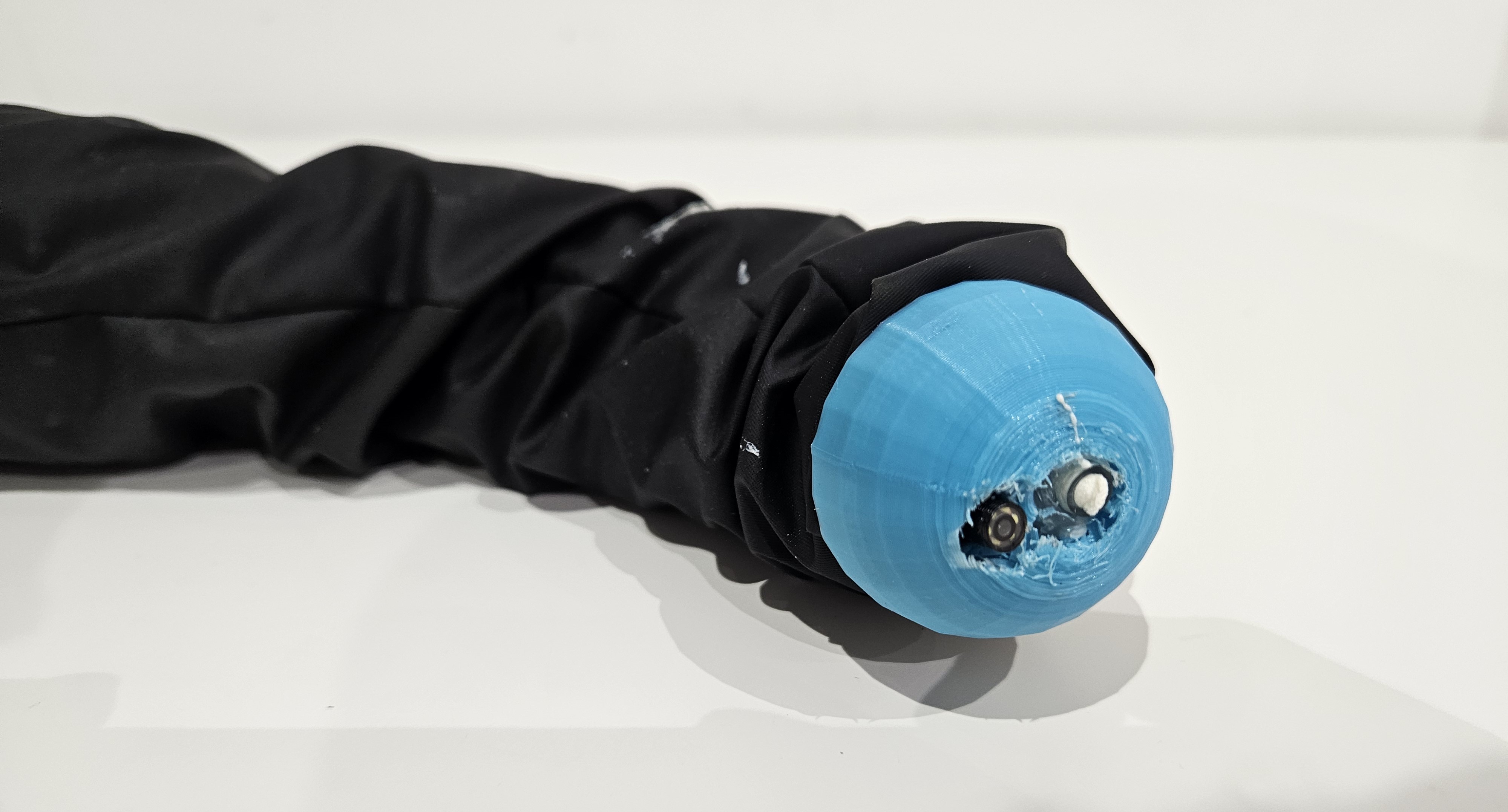}

  \caption{Front view of the hybrid robot showing its sprayer and camera integrated in a hard cap.}
\end{figure}

There is a high demand for robots that can navigate hard-to-access spaces. The biggest challenges, for example, in relation to  inspection of pipe networks are navigating within the pipe network \cite{c12} and delivering a payload  to a distant target area. Current robots are limited by their ability to navigate these environments due to their lack of mobility \cite{c13}. Generic track operated robots are often ineffective as they are only able to navigate on level surfaces. As a result, they can get stuck easily and are difficult to recover. Drones may be viable in some instances \cite{c14} but they too have their own disadvantages as they overly disturb the environment in which they travel through,  are more difficult to control and require a stable connection with the operator which further adds to their operational complexity.

There are only a handful of papers that look at active steering in eversion robots, leaving a relatively empty playing field, ripe for novel methods \cite{c15}, \cite{c16}.  

In this work  a new idea is introduced that combines a soft eversion robot with a continuum robot at its tip as shown in figure \ref{fig1}, to create a hybrid robot whose movements are precise and responsive while retaining soft, flexible, and compliant characteristics. This approach offers several advantages  over traditional rigid or flexible robots. It enables the robot to navigate through complex environments with greater ease, as it can choose a path to traverse whilst being everted. This is in contrast to most traditional eversion robots which require pneumatically inflated pouches or actuators to be sewn onto them in order to navigate certain elements of their environment.  Our hybrid robots can be turned using a joystick which controls a decoupled steering mechanism at the tip, that allows the user to actively change its path while in a pipe or enclosed space. Altering the direction of the tip of the robot and maintaining the pressure within the soft robot body enables it to move forward in specified directions. Unlike with cPAMs \cite{c15} or with integrated pneumatic pouches, this method does not enable the robot to hold its shape once it has been steered and is therefore reliant on its environment to do that. This steering capability however is crucial, especially at T-junctions, at which point the end user can guide the robot. Extending the capabilities of this hybrid robot, it enables the transportation of various payloads, including sensors, liquids, and aerosols. The decoupled steering mechanism at the tip facilitates precise control over the robot's movements, making it a versatile tool for applications requiring controlled transportation in challenging and dynamic environments and provides a solid mounting point for sensors and payloads that need to be delivered.

\section{Related Works}
Several steering methods for eversion robots have been proposed by researchers, including those that involve internal steering devices and others that use pneumatic actuators. One idea proposed is that of asymmetric lengthening of the tip which causes extension of either side to induce the robot’s turning behaviour. This could be controlled by several stimuli such as the use of a camera and a light source to steer the robot towards the light \cite{c4}. These methods however cannot be sustained for longer lengths and do not provide any form of active steering after several bends.

Another steering mechanism used pouches that were sewn into the side walls of the eversion robot, allowing predetermined steering at specific stages along its path. This meant that actuators caused the inflated backbone of the eversion robot to bend in different directions \cite{c16}. This was later revised by Kübler et al’s cylindrical pneumatic artificial muscle (cPAM) design \cite{c17} and then further developed by Kübler et al’s selective steering cPAM design \cite{c15}. 

There are many advantages of using such methods to control eversion robots - specifically, in the case of  pneumatic pouches and cPAM, its ability to increase the angle to which the eversion robot is able to bend - as demonstrated by the 133\% improvement in bending angle per Abrar et al’s paper \cite{c16}. However, some of the most significant disadvantages of using pneumatic pouches and cPAM’s are the added complexity of needing extra air supplies, valves and regulators, the limited actuation range, and the potential multiple failure points due to the number of stitched regions, which make the end-product bulky, less manoeuvrable and less reliable. Relying on predetermined pouches presents challenges in many nuclear decommissioning scenarios as there is often a lack of a prior knowledge of legacy pipe structures that makes identifying specific locations for bends and turns along the path length challenging. A method of active steering must therefore be developed to enable directed steering as the robot traverses along a pipe network.

\begin{figure}[t]
  \centering
  \includegraphics[width=1\columnwidth]{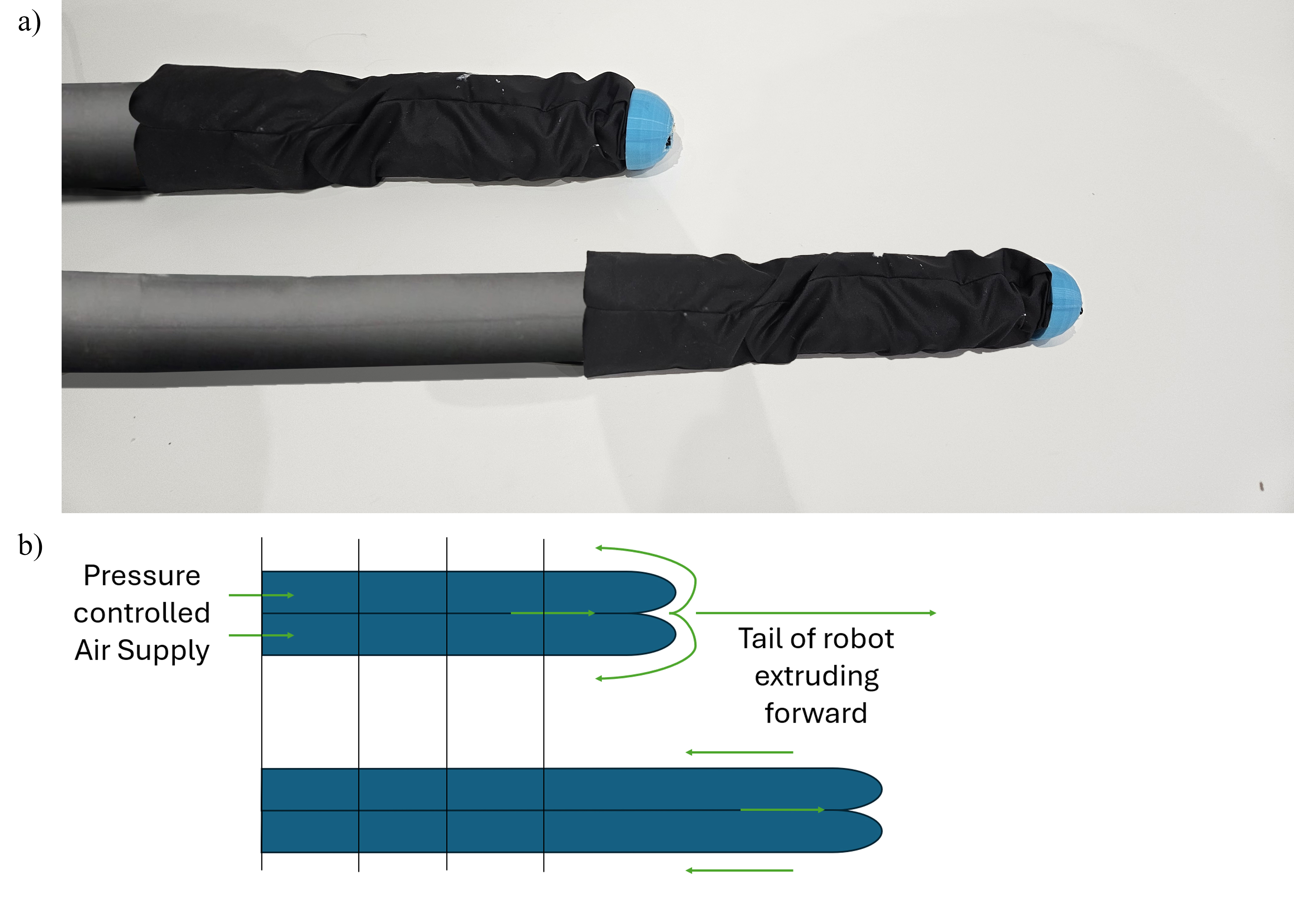}
  \caption{Illustration depicting the eversion robot in action in figure a), showcasing its operational mechanism and dynamic expansion process. Figure b) illustrates the operation of how tail extrusion causes no shift in the outer lining of the robot.}
\end{figure}
\newpage

\section{Contributions}
The contributions of this work are:
\begin{itemize}
    \item A three degrees-of-freedom eversion-continuum hybrid robot that uses a series of servos to allow selective steering of an eversion robot tip.
    \item A method by which the robot can deliver sensors to remote areas and facilitate the controlled application or spraying of liquids and aerosols in these remote areas, showcasing the robot's potential in decontaminating remote nuclear-contaminated vessels and glove-boxes.
\end{itemize}

\begin{figure}[t]
  \centering
  \includegraphics[width=1\columnwidth]{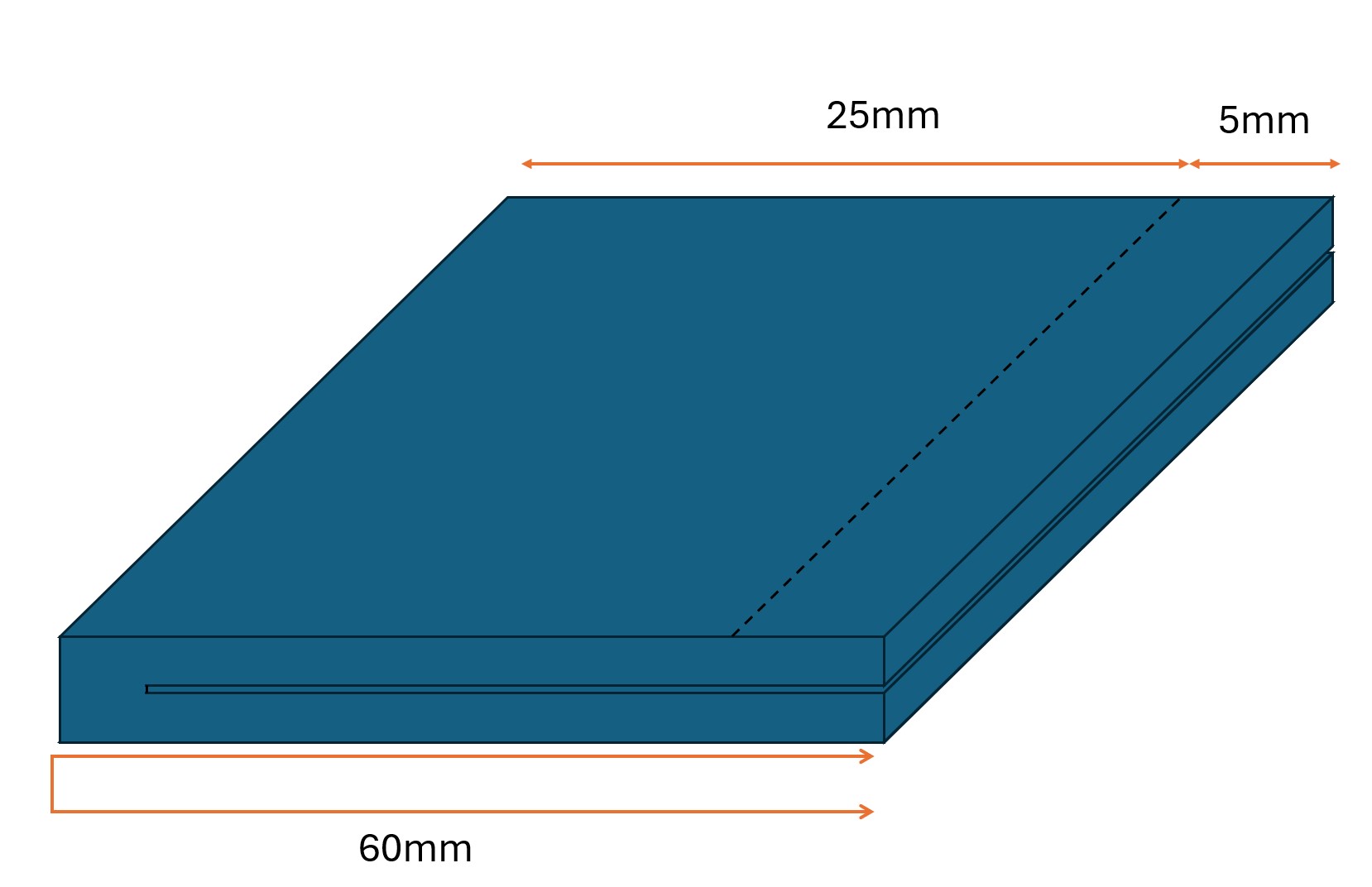}
  \caption{60 mm Nylon ripstop material folded in half and stitched 5 mm from the edge. The dotted line shows where the seam should be.}
\end{figure}
\begin{figure}[b]
  \centering
  \includegraphics[width=1\columnwidth]{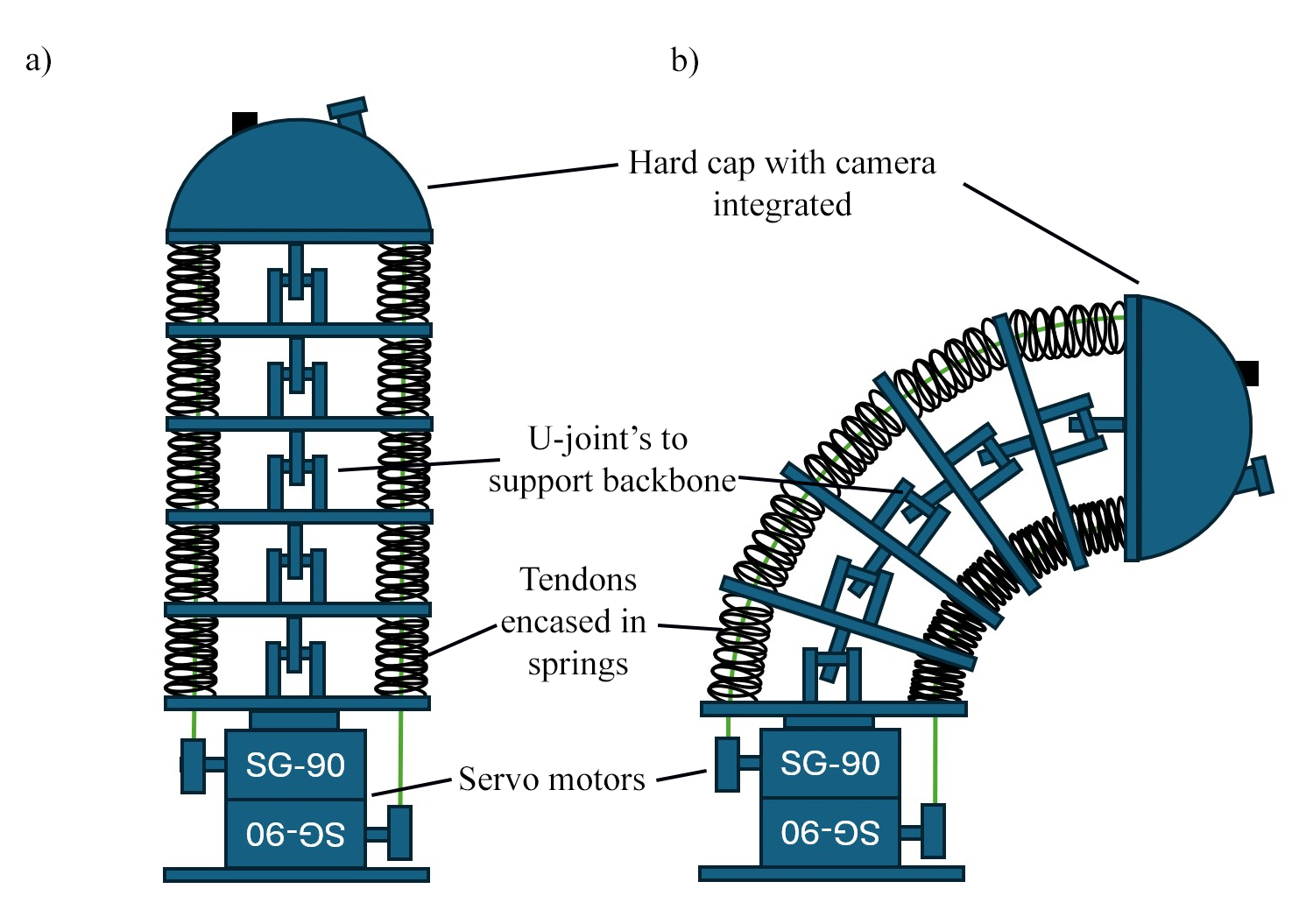}
  \caption{Detailed view of the tip structure: Illustrating how the robot turns when the servos are not actuated a) and actuated b).}
\end{figure}

\section{Design}

\subsection{Eversion Robot design}
The ‘vine’ section of the robot is made of nylon ripstop material. The material is cut into strips with a width of around 60 mm. The material is folded over along the short edge and made to overlap. A seam line is drawn on the material and a guide setup to measure 5 mm from the edge. The strips are then sown together forming a seam. This is done using a Singer sewing machine with nylon thread. The seam is then covered with vinyl that is heat pressed to create an airtight seal. This is then tested by inflating the robot and inspecting for leaks. Once this is done, the end is tied off and the robot inverted. A 5 mm airline tube is inserted and sealed to the tail to allow it to inflate.

\subsection{Cap Design}
To enable the tip of the eversion robot to move in several distinct directions, we attached a continuum robot to its tip. Consequently, the robot can turn in both the X and Y axis providing 2 degrees of movement –  effectively introducing selective steering into the robot’s repertoire. The body of the continuum section is made of 3D printed discs connected in the centre by a u-joint, and by springs along the edges. Each segment of the continuum section has 4 springs which are set into raised 3D printed pedestals that stop them from sliding off the edge of the discs. The central u-joint provides a solid backbone to the tip which means that it cannot be compressed if the robot hits a wall. It also allows the robot to be controlled more effectively as the desired angle can be determined given the fixed distance between discs. Each axis of springs coils around a nylon wire that spans the length of the robot to a servo wheel. Each of the 4 servos is responsible for its own quadrant - their combined effect determining the direction of the robot.

\begin{figure}[t]
  \centering
  \includegraphics[width=1\columnwidth]{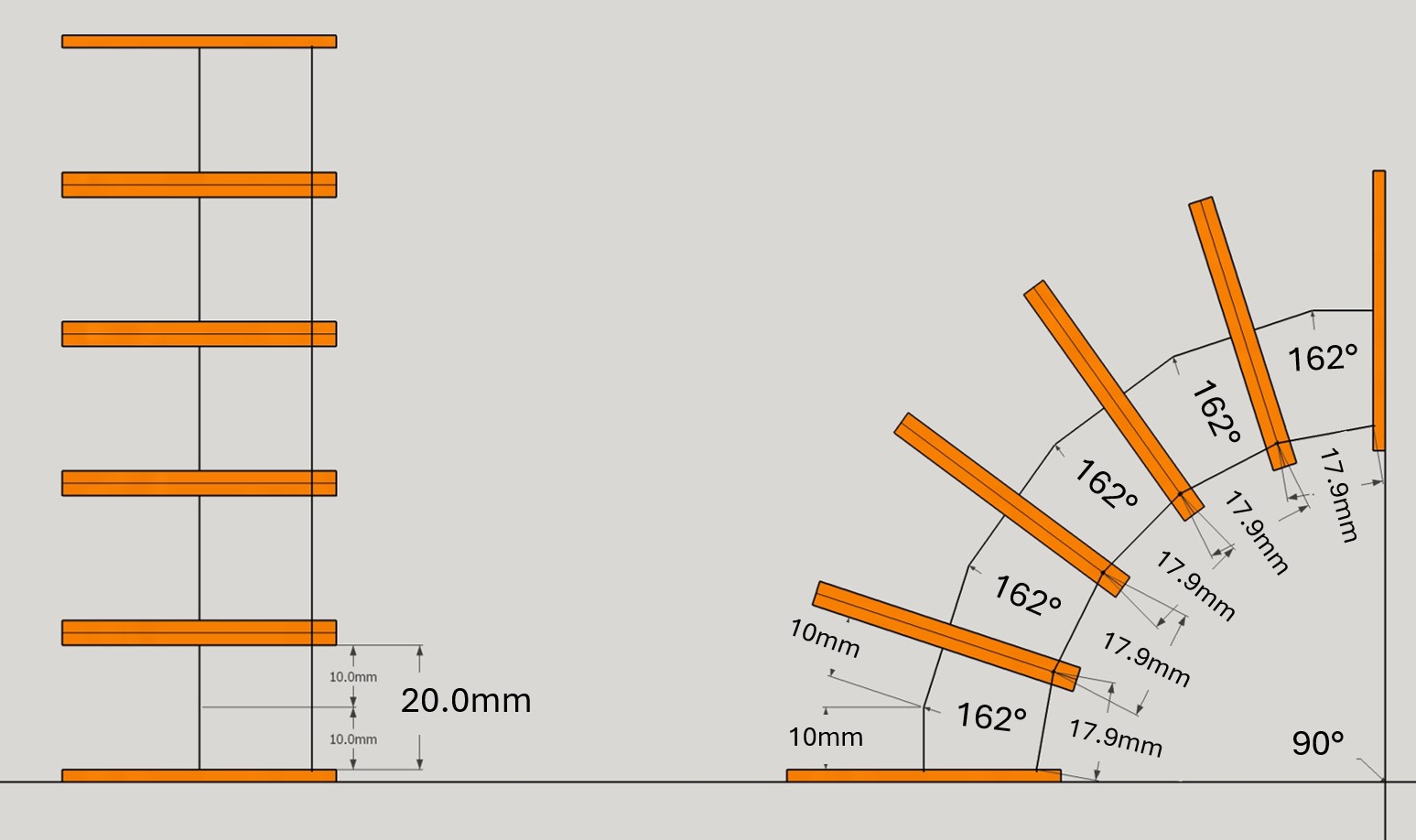}
  \caption{CAD model demonstrating theoretical bending angle and length of nylon wire when the robot is turned 90 degrees.}
  \label{cad}
\end{figure}

\subsection{Modelling the angles in CAD}
Using CAD modelling as shown in figure \ref{cad}, it is possible to measure the length of nylon thread from the base of the robot to the tip, both when the robot is straight and when it is turned. The difference in length can then be calculated. Calculating the angle between each disc was achieved by dividing the angle (90 degrees) by the number of plates minus 1, in this case 90/5=18 degrees. By modelling this, it was possible to determine that the arc length of the eversion robot on the compressed side is 85.5 mm whereas the uncompressed length is 100 mm. This meant that the motors used would require enough torque and angle to pull the nylon wire by 14.5 mm. It was assumed that this is also the distance that each spring would be compressed by.

\subsection{Computing the forces required to actuate the tip structure}
To work out the maximum force that would be required by each spring it was assumed that the maximum length of compression was not 2.9 mm but rather 5 mm which provided a 72\% tolerance over the required length to account for any discrepencies in the rated torque figures provided in the datasheets.
Based on the 5 mm compression, the forces required to compress each spring were subsequently modelled. 20 mm long 304 grade stainless steel springs, 6mm in diameter and made of 0.05 mm wire, were used in this application.
Using the following equation, the spring stiffness could be determined:
\begin{equation}
k=\frac{G \times d^4}{8 \times N \times D^3}
\label{kequation}
\end{equation}
where $G$ is the shear modulus of the material, $d$ is the diameter of the wire (in metres), $N$ is the number of active coils and $D$ is the mean coil diameter.
The shear modulus of the material was calculated using the Young’s modulus and Poisson’s ratio of 304 stainless steel, which are approximated as 190 GPa and 0.27 respectively. Using this formula:
\begin{equation}
G=\frac{E}{2(1+v)}
\label{shearmodulus}
\end{equation}
where $E$ is Young’s modulus and $v$ is the Poisson’s ratio, and substituting for the above values, a shear modulus value of 74.8 GPa was obtained, which was rounded to 75 GPa.
Using \eqref{shearmodulus}, and a shear modulus of 75 GPa, a diameter of 0.05 mm, 6 active coils and a mean coil diameter of 5.5 mm, which was determined by subtracting the outer diameter of the spring by the wire diameter, the spring constant was calculated to be approximately 586 N/m.
Using Hooke’s law, it is possible to approximate the force exerted by each spring, given that each spring is required to move 5 mm and has a spring constant of 586 N/m.
\begin{equation}
F=-k x=-587 \times-0.005=2.935
\end{equation}
Where $k$ is the spring constant and $x$ is the length in metres.
This meant that the force required to compress each spring was approximately 2.9 N. It was then possible to calculate the torque required by the motor to successfully pull the nylon wire and compress the springs by the required distance, which will be further examined in the following section.

\subsection{Choosing the right motors}
With such a small form factor required to successfully fit into a 2-inch diameter pipe, cutting down on the size of the rigid structure is the most significant challenge. The eversion robot has a unique feature of being able to traverse through piping of variable diameter, as the soft compliant material on the outside can expand and contract accordingly to fit the size of the pipe. This however is not possible when rigid material is used at the tip of the eversion robot such as a 3D printed structure – indeed the maximum diameter of pipe that the eversion robot is able to travel along would be determined by the maximum width of the solid structure within the tip. As a result, reducing the diameter of the steering mechanism is the highest priority. However, one must also carefully consider the size of the motor in relation to the force it can exert, especially given the small form factor constraints and the 45 mm disc diameter. Another key requirement of the motor is to be capable of providing its rotational position to inform the operator of the robot’s tip position. This requirement rules out the use of direct drive DC motors, leaving stepper and servo motors. 

Stepper motors are synchronous DC motors that do not spin continuously but rather use a series of coils arranged in phases. These can be turned on or off in quick succession to turn the rotor shaft by a fraction of a rotation at a time, each one called a step. Suitably small stepper motors do exist, with models such as the NEMA 6 stepper motor with a highly stackable body size of just 15 mm x 15 mm x 30 mm. However, stepper motors have limited torque and the NEMA 6, for example, is rated to 0.58 Ncm (0.0000058 kg/cm). 

Having eliminated stepper motors, servo motors were identified as the most appropriate option. Although limited in their motion range and not as small as the NEMA 6 stepper, servo motors represent a good compromise between DC motors and steppers as they offer high torque in relation to their size due to their gear ratio, and can transmit the current position of the shaft through built-in encoders. They do however come at a cost in terms of size, with the smallest production servos, the SG90/MG90 family, having dimensions of 22.8 mm x 12.2 mm x 28.5 mm. Because of this, the servos must be stacked on top of each other and offset by 90 degrees to each other to allow each one to control its own nylon wire. Different manufacturers of MG90 servos were examined, as well as different variants of servo motors, to identify the most appropriate option. The most important features of the servo to consider are their operating angle and stall torques. Using the servo with the largest operating angle allows the design to be smaller, as the servo horn, a cylindrical disc that houses the nylon spool, has a smaller diameter. However, this negatively impacts the servo's ability to pull the wire as the torque decreases.
The torque diminishes as the distance from the centre of rotation, where the force is applied, decreases. This is due to torque being the product of force and distance from the axis of rotation as shown by this equation:
\begin{equation}
\tau=F \times r
\end{equation}
where $\tau$ is the torque, $F$ is the force applied and $r$ is the distance from the centre of rotation to the point where the force is applied. This means that a compromise must be made between the torque that the servo is able to provide and the operating angle.
The servos examined in this study are listed in Table \ref{Servo}.

\begin{table}[b]
\caption{Servo Torque Performance Comparison of different models at varying voltages.}
\label{Servo}
\begin{center}
\begin{tabular}{|l|c|c|c|}
\hline
\textit{\textbf{Servo}} & \textbf{\begin{tabular}[c]{@{}c@{}}Operating \\ Angle\end{tabular}} & \textbf{\begin{tabular}[c]{@{}c@{}}Torque at \\ 4.8 V (kg/cm)\end{tabular}} & \textbf{\begin{tabular}[c]{@{}c@{}}Torque at \\ 6 V (kg/cm)\end{tabular}} \\ \hline
\textbf{SG90}           & 180                                                                 & 1.2                                                                         & 1.6                                                                       \\ \hline
\textbf{MG90s}          & 180                                                                 & 1.8                                                                         & 2.2                                                                       \\ \hline
\textbf{DMS-MG90-A}     & 270                                                                 & 1.3                                                                         & 1.5                                                                       \\ \hline
\textbf{DS-S006L}       & 300                                                                 & 1.0                                                                         & 1.2                                                                       \\ \hline
\rowcolor[HTML]{B1F87A} 
\textbf{DM-S0090MD}     & 270                                                                 & 1.8                                                                         & 2.0                                                                       \\ \hline
\end{tabular}
\end{center}
\end{table}

The two candidate servos considered were the DS-S006L and the DM-S0090MD on account of their increased operating angle. The question remained as to whether the torque provided would suffice to compress the springs.
This was determined by multiplying 2.95 N by 5 to find the total axial force required to compress all 5 springs, which is 14.75 N. Connecting this to a spool of radius 12.50 mm the required torque was determined by:
\begin{equation}
\begin{gathered}
\quad \text { Required torque }=\text { Force } \times \text { Effective Radius } \\
=14.75 \mathrm{~N} \times 0.0125 \mathrm{~m}=0.184 \mathrm{Nm} \\
\text { Stall torque } D S-S 006 \mathrm{~L}(\mathrm{Nm}) \\
=1.2 \mathrm{~kg} / \mathrm{cm} \times 0.01 \mathrm{~m} / \mathrm{cm} \times 9.81 \mathrm{~m} / \mathrm{s}^2=0.1177 \mathrm{Nm} \\
\text { Stall torque } D M-S 0090 \mathrm{MD}(\mathrm{Nm}) \\
=2.0 \mathrm{~kg} / \mathrm{cm} \times 0.01 \mathrm{~m} / \mathrm{cm} \times 9.81 \mathrm{~m} / \mathrm{s}^2=0.1962 \mathrm{Nm} \\
\text { Stall torque } D M-S 0090 \mathrm{MD}>\text { Required Torque }
\end{gathered}
\end{equation}
The torque requirement exceeds the stall torque of the DS-S006L but is within the stall torque of S0090MD. Therefore, S0090MD with a stall torque of 2.0 kg/cm at 6 volts is deemed suitable for pulling the wire and compressing the spring under the given conditions.
However, it is also necessary to determine if enough wire would be pulled to compress all the springs by approximately 15 mm, taking into account the size of the servo hook and the angle at which the servo is used.
Calculating the arc length of a circle with a radius of 12.5 mm and an arc angle of 270 degrees:
\begin{equation}
s=\frac{270}{360} \times 2 \pi \times 12.5 \mathrm{~mm} \approx 58.9 \mathrm{~mm}
\end{equation}
This confirms that the candidate servo motor would be able to apply sufficient pulling force onto the wire to fully compress the continuum robot before reaching the servo motor's limits.

\section{Experimental Evaluation}
The robot is controlled using a control panel, facilitating operation through buttons and joystick's connected to an Arduino uno. The control signals are then passed down a tether to the tip of the robot, controlling the servos and the axial movement of the hybrid robot. The Arduino is also connected to a SMC ITV2050 pneumatic regulator that in turn is connected to a compressor supplying pressure to the robot. This allows the operator to control movement in both sections of the hybrid robot from a single control panel. 

The performance of the hybrid robot was tested by conducting several real-world experiments to observe its effectiveness and limits. Attention was paid to the ability of the robot to successfully deliver a set of sensors/camera to a remote location and to then deploy a “decontaminating” agent or sprayable expanding foam to the target area. 

This ability would pave the way for its use in a variety of nuclear industry applications – most obviously those involving remote access and decontamination of hazardous material.  The precise workings of the continuum robot are key if they are to allow for effective decontamination. Most importantly, the integration of the control panel enables human-safe remote operation, with the operator being able to be based some distance away, or even in an entirely separate location.  This is clearly useful in relation to areas that have toxic vapours or concerning levels of radiation.

\begin{figure}[t]
  \centering
  \includegraphics[width=1\columnwidth]{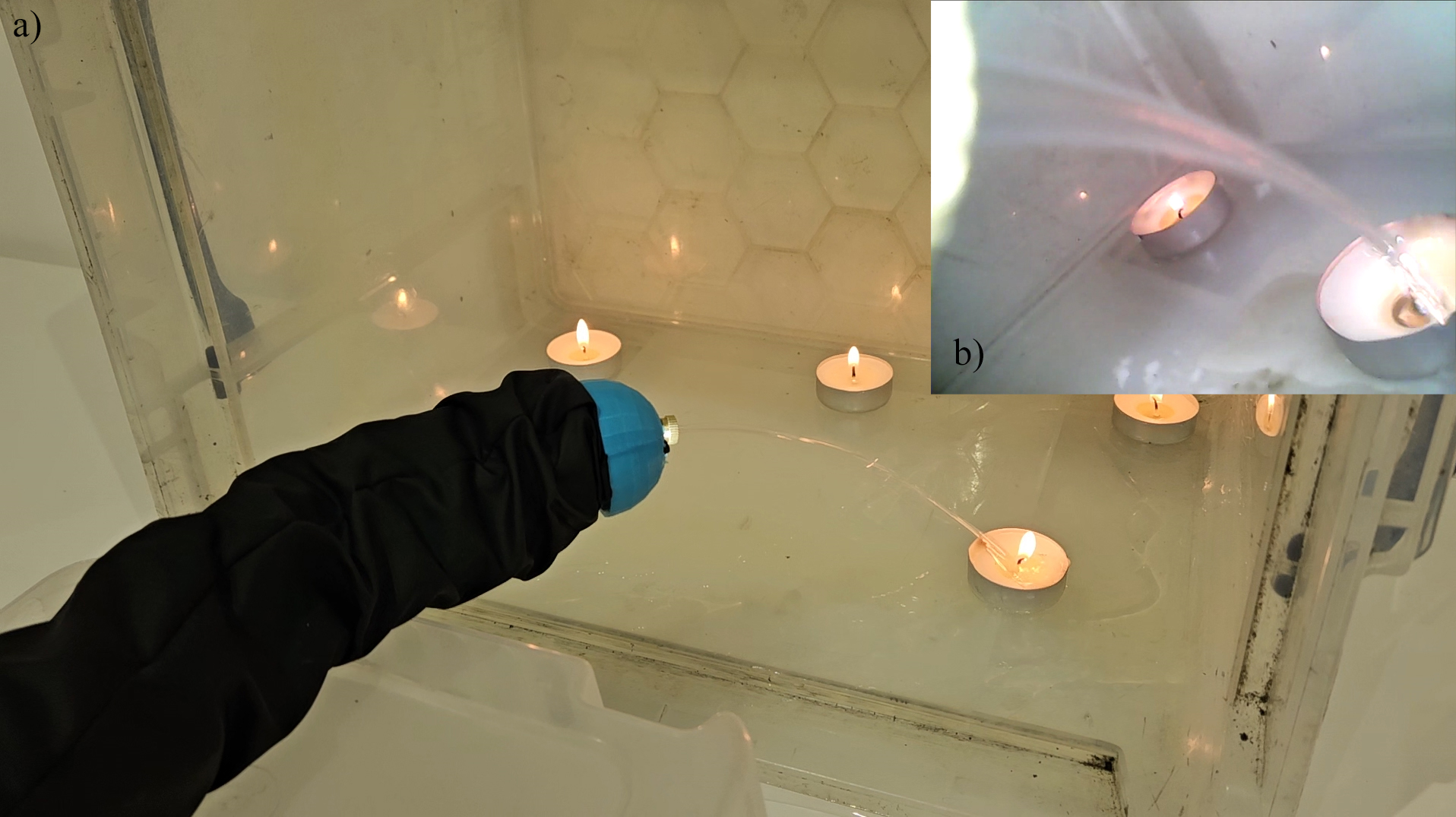}
  \caption{a) Hybrid robot demonstrating its versatility in simulating scenario 1, effectively extinguishing a candle by employing a water-spraying mechanism. b) A POV view of the target, using the built in tip mounted micro camera.}
  \label{pov}
\end{figure}

\begin{figure}[t]
  \centering
  \includegraphics[width=1\columnwidth]{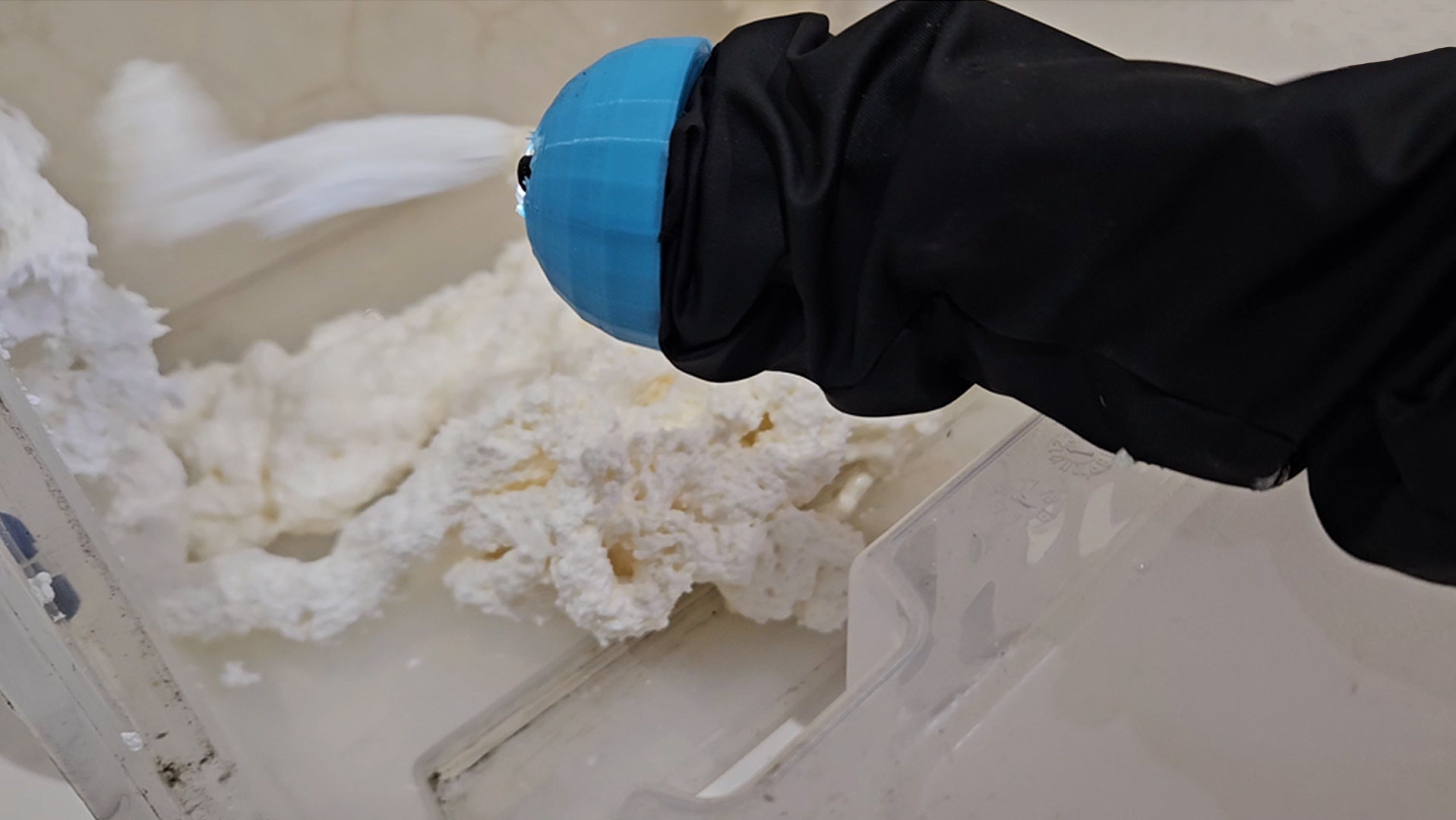}
  \caption{Hybrid robot showcasing adaptability in simulating scenario 3, employing an expanding foam application to effectively cover the walls of the glove box.}
  \label{spraypic}
\end{figure}

The three test scenarios conducted were tailored to the liquids that were sprayed:
\begin{itemize}
    \item 
    For the liquid decontaminating agent, a course through a pipe was set up for the eversion robot to traverse, at the end of which it had to identify the problem area, as in the case shown in Figure \ref{pov}, and then deploy the agent (water) to remedy the issue within a model glove box. 
    \item 
    For the second test, we  wanted to test the robot’s ability to precision spray in a specified direction. For this a 6x10 grid was produced on A3 cold press paper (watercolour paper) which has a high water absorbency rate. This was mounted on the inside wall of one of the model glove boxes and used as a target onto which the eversion robot had to spray aerosol paint. The robot’s ability to individually spray each of the squares was tested within a 5-minute time span. The results were recorded and the experiment repeated 10 times to find the percentage of paint coverage on the piece of paper.
    \item 
    The third scenario is like the first, though without any need to extinguish fire. In this scenario the robot is used to immobilize a certain object or material. The eversion robot traverses the pipe into a remote vessel or glove-box and is then required to spray material in all directions to successfully treat all surfaces of the box, as shown in figure \ref{spraypic}. 
    
\end{itemize}

\begin{table}[b]
\caption{Success Percentage of Spraying Grids using Hybrid Robot}
\label{spray}
\begin{center}
\begin{tabular}{|l|c|c|}
\hline
\textbf{Test number} & \textbf{\begin{tabular}[c]{@{}c@{}}Number of \\ squares\end{tabular}} & \textbf{\begin{tabular}[c]{@{}c@{}}Percentage of \\ grid sprayed\end{tabular}} \\ \hline
\textbf{1}           & 60                                                                    & 100                                                                            \\ \hline
\textbf{2}           & 59                                                                    & 98.3                                                                           \\ \hline
\textbf{3}           & 57                                                                    & 95                                                                             \\ \hline
\textbf{4}           & 60                                                                    & 100                                                                            \\ \hline
\textbf{5}           & 56                                                                    & 93.3                                                                           \\ \hline
\textbf{6}           & 59                                                                    & 98.3                                                                           \\ \hline
\textbf{7}           & 59                                                                    & 98.3                                                                           \\ \hline
\textbf{8}           & 55                                                                    & 91.7                                                                           \\ \hline
\textbf{9}           & 60                                                                    & 100                                                                            \\ \hline
\textbf{10}          & 57                                                                    & 95                                                                             \\ \hline
\textbf{Average}     & \textbf{58.2}                                                         & \textbf{96.99}                                                                 \\ \hline
\end{tabular}
\end{center}
\end{table}

These experiments verified that the robot’s ability to spray water, paint and foam in specified directions is feasible, in each case the robot achieving its intended purpose. The eversion robot was able to successfully lead the continuum robot to the required area whereupon it successfully deployed the water, paint and the foam. 
The results of the grid spray test are shown in Table \ref{spray}, where it is observed that the robot achieved more than a 95\% success rate. Upon closer examination of the data, it becomes evident that the 5\% lack of spray in the empty squares was randomly distributed across the 10 experiments. This randomness suggests that limitations in the mechanics of the system, combined with the inherent difficulty of operator control, contributed to the observed shortcomings.

The uneven distribution of the spray pattern may be attributed to mechanical factors such as variations in pressure. Additionally, operator control plays a crucial role in the effectiveness of the system. 
However, this proof of concept successfully demonstrates a technology readiness level of 4, with a prototype successfully demonstrated in a lab environment.  

\section{Conclusion}
We present a novel hybrid continuum-eversion robot that can grow and decontaminate pipe structures and access hard to reach remote areas. The user can operate the robot using a controller that commands a set of servos. These servos control a continuum robot, steering its tip. Simultaneously, a pneumatic soft eversion robot moves the robot forward, creating a three degrees-of-freedom hybrid robot. This hybrid robot is capable of transporting liquids, aerosols or sensors to remote areas on account of its modular tip that can be interchanged between a cone structure, to transport a camera and sprayer, and a platform structure to transport a sensor payload. 

Possible applications include the exploration of confined spaces such as air conditioning ducting, navigation and sensing of underground ducting, navigation and decontamination of glove-boxes, generic piping, enclosed vessels, water tanks and more. This hybrid robot shows significant promise in relation to the nuclear industry.

There were some limitations with the robot, mainly due to the manufacturing processes involved and the resulting air leakage. Future work will need to focus on better sealing methods, such as heat sealing or ultrasonic sealing of special TPU coated nylon ripstop material.
Another limitation was that the cap that held the soft and hard components of the robot was only friction fit. This caused some problems when everting at different angles, especially if there was a vertical drop. This would sometimes cause the continuum robot section of the robot to slide off and end up irretrievable other than by using the tether to pull it back. This suggests that future considerations should address how to better retract the robot safely. 
Further research considerations include investigating the feasibility of decreasing the footprint of the robot to enable deployment in smaller diameter pipes. Currently the most significant restriction is that the robot can only work within piping that has a diameter greater than the maximum width of the hard PLA material in the continuum robot.

Future research directions also include extending the design to use a camera with AI/computer vision components to guide the robot towards or away from certain stimuli. This, along with gyroscopic tracking of the tip of the robot could also be used to create a 3D visualisation of the area that the robot is spraying, providing real-time feedback on progress to the operator. The gyroscopic tracking of the head could also be used to create a visual representation of the robot itself, providing an estimation of its everted length, the path it has taken and the angle at which the robot is pointing. A  live camera feed would also facilitate remote steering. A study on the effects of radiation on the materials that are used should also be carried out to test the viability of the solution for the intended purposes and environments. 

\section*{Acknowledgments}
The authors thank Mish Toszeghi for his comments and review. The authors thank Abu Bakar Dawood for his valuable help.

\bibliographystyle{unsrt}

\end{document}